%% file: arxiv.tex
\documentclass[10pt,twocolumn,letterpaper]{article}

\usepackage[pagenumbers]{cvpr} 

\usepackage{times}
\usepackage{epsfig}
\usepackage{graphicx}
\usepackage{amsmath}
\usepackage{amssymb}
\usepackage{float}
\usepackage{caption}

\usepackage{booktabs}

\usepackage{bm}
\usepackage{amsfonts}
\usepackage{algorithmic}
\usepackage{textcomp}
\usepackage{colortbl}

\usepackage{algorithm}
\usepackage{pifont} 
\usepackage{bbding} 
\usepackage{bbm}
\usepackage{indentfirst}

\usepackage{multirow}
\usepackage[T1]{fontenc}
\usepackage{color}

\usepackage[breaklinks=true,bookmarks=false]{hyperref}

\input{preamble}
\title{McSc: Motion-Corrective Preference Alignment for Video Generation with Self-Critic Hierarchical Reasoning}

\author{
    Qiushi Yang  \quad
    Yingjie Chen \quad
    Yuan Yao     \quad
    Yifang Men   \quad
    Huaizhuo Liu \quad
    Miaomiao Cui \\[5pt]
    Tongyi Lab, Alibaba Group \\
    Project webpage: https://qiushiyang.github.io/McSc/ \\
}

\begin{document}
\maketitle
\input{sec/0_abstract}    
\input{sec/1_intro}
\input{sec/2_related}
\input{sec/3_preliminary}
\input{sec/4_method}
\input{sec/5_exp}
\input{sec/6_conclusion}
{
    \small
    \bibliographystyle{ieeenat_fullname}
    \bibliography{main}
}


\end{document}

%% file: sec/0_abstract.tex
\begin{abstract}
Text-to-video (T2V) generation has achieved remarkable progress in producing high-quality videos aligned with textual prompts. However, aligning synthesized videos with nuanced human preference remains challenging due to the subjective and multifaceted nature of human judgment.
Existing video preference alignment methods rely on costly human annotations or utilize proxy metrics to predict preference, which lacks the understanding of human preference logic. 
Moreover, they usually directly align T2V models with the overall preference distribution, ignoring potential conflict dimensions like motion dynamics and visual quality, which may bias models towards low-motion content. 
To address these issues, we present Motion-corrective alignment with Self-critic hierarchical Reasoning (McSc), a three-stage reinforcement learning framework for robust preference modeling and alignment. Firstly, Self-critic Dimensional Reasoning (ScDR) trains a generative reward model (RM) to decompose preferences into per-dimension assessments, using self-critic reasoning chains for reliable learning. 
Secondly, to achieve holistic video comparison, we introduce Hierarchical Comparative Reasoning (HCR) for structural multi-dimensional reasoning with hierarchical reward supervision. 
Finally, using RM-preferred videos, we propose Motion-corrective Direct Preference Optimization (McDPO) to optimize T2V models, while dynamically re-weighting alignment objective to mitigate bias towards low-motion content. Experiments show that McSc achieves superior performance in human preference alignment and generates videos with high-motion dynamic.
\end{abstract}

%% file: sec/1_intro.tex
\section{Introduction}
\label{sec:intro}

\begin{figure}[h!]
\centering
\includegraphics[width=0.49\textwidth]{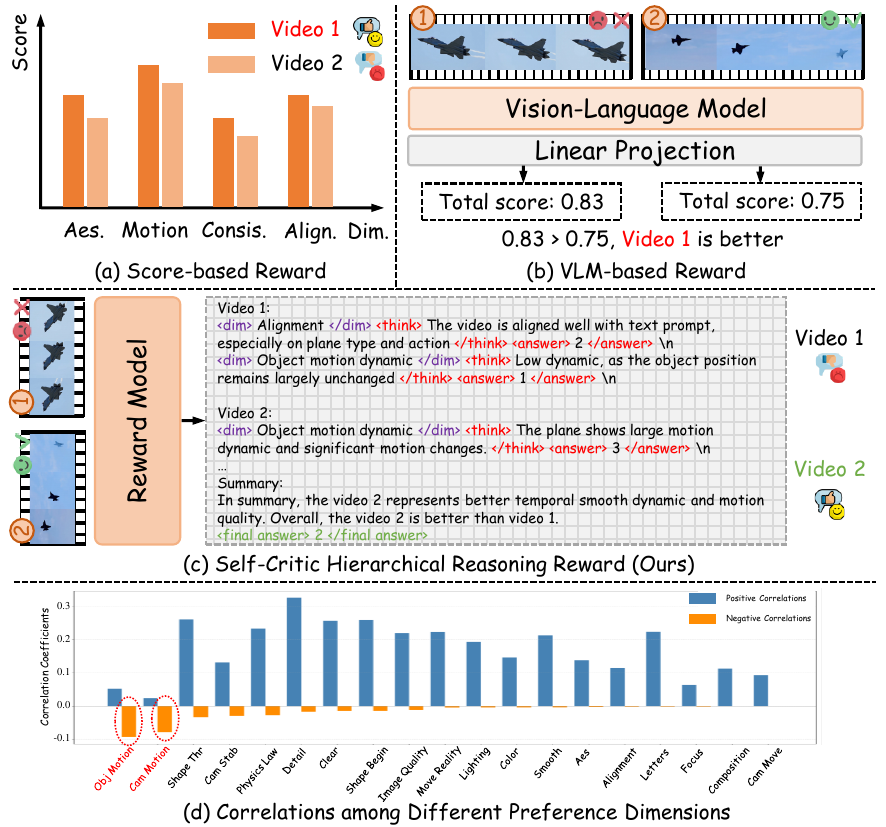} 
\caption{
Given two videos where video 2 is preferred by humans, (a) and (b) show score-based and VLM-based reward approaches failing to make reliable judgments. (c) In contrast, our self-critic hierarchical reasoning correctly identifies the preference via single-dimension then holistic reasoning. (d) Dimension correlations reveal a strong negative coupling between motion dynamic and visual quality, indicating a potential source of reward bias.}
\label{fig-motivation}
\end{figure}

Recent advancements in text-to-video (T2V) generation~\cite{vc2,wan,Cogvideo,Cogvideox,Hunyuanvideo,Perception-as-Control,Follow-your-click,Open-sora} have significantly improved the ability on producing high quality, temporally coherent videos that align well with input textual prompts. 
However, despite these improvements, effectively aligning synthesis videos with real human preference remains a critical challenge since the preference of video content is inherently subjective and multifaceted. Traditional approaches rely on customized metrics, which are hard to capture holistic nature of human judgment, leading to suboptimal alignment with human expectations.

\begin{figure*}[t]
\centering
\includegraphics[width=0.99\textwidth]{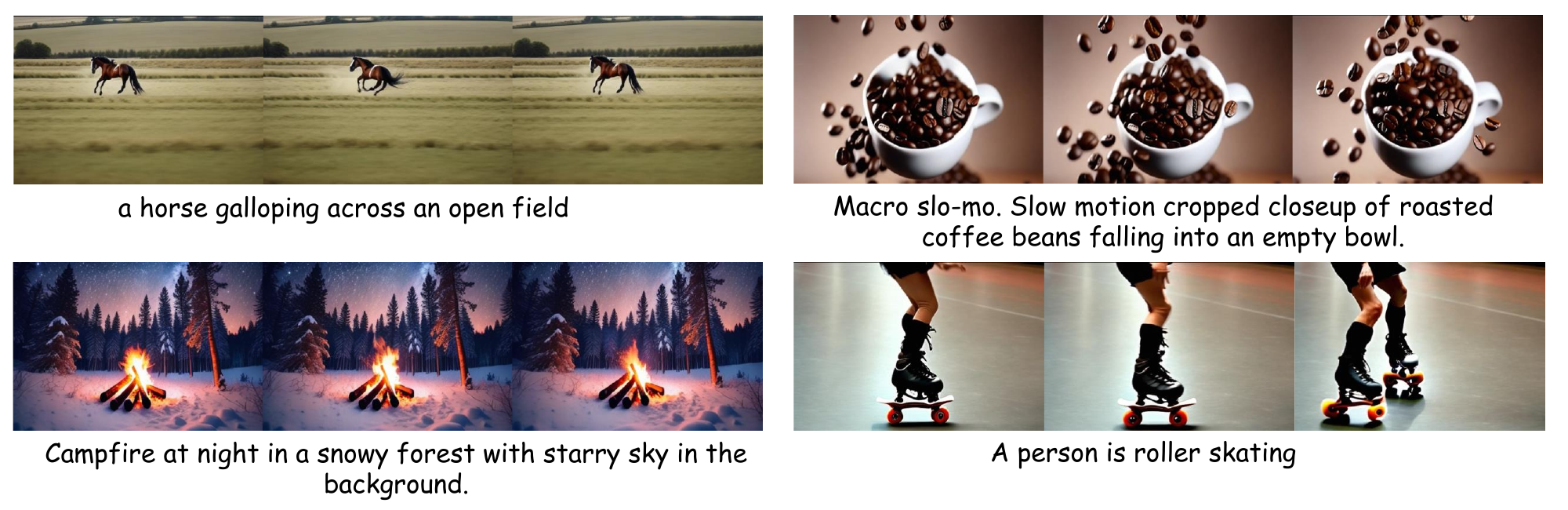}
\caption{Alignment video generation results of the proposed method from the text-to-video model.}
\label{fig-visualization1}
\end{figure*}

To address this misalignment, recent works~\cite{videoalignment,VideoDPO,Mj-video,ImproveFeedback,HuViDPO} explores reinforcement learning (RL) to align video generation models with human preference via a \textit{preference prediction then alignment} pipeline.
For \textit{preference prediction}, many of them utilize off-the-shelf models or heuristic operations~\cite{clip,VideoDPO} to calculate proxy metrics, such as CLIP score or aesthetic score. Although informative, they are limited by the subjectivity and \textbf{lack an understanding of the intrinsic judgment logic} behind human preference criteria, incurring unreliable estimation.
For example, a video may outperform another on proxy metrics yet be less preferred by humans (see Figure~\ref{fig-motivation} (a)). 
Some recent studies~\cite{Visionreward,unifiedreward} employ vision-language models (VLMs) with a linear projection to yield an overall quality score, while they regress to scalar scores without explicit multi-dimensional reasoning, limiting the generalization and pairwise video comparisons (see Figure~\ref{fig-motivation} (b)).
Furthermore, for \textit{preference alignment}, video generation models are typically optimized to match the overall preference distribution. However, human preferences involve diverse, sometimes conflicting dimensions. Our analysis in Figure~\ref{fig-motivation} (d) reveals a \textbf{strong negative correlation between temporal motion dynamics and other static visual quality}. For example, low-motion videos tend to score higher on visual quality aspects. Hence, direct alignment with such biased distributions may cause reward hacking, where the video generation models favor low-motion but high overall metric outputs, reducing motion dynamic (see Table~\ref{table-ablation}).

We solve these problems by deconstructing the real human video preference judgment process, which involves first evaluating videos across multiple quality dimensions and then integrating these comparisons into a holistic preference. 
Inspired by this, we design a \textit{generative reward model} that mimics this two-step reasoning: first, it \textbf{learn the latent criteria underlying each human judgment dimension}; second, it \textbf{capture the holistic reasoning logic behind overall preference}. This enables the reward model to accurate estimate preference and construct high-quality positive-negative video pairs.
Using these pairs, we align the generative model with the preference distribution. To mitigate motion-related alignment bias in existing approaches, we encourage the video generation model to \textbf{favor videos with higher motion dynamics}, even if other qualities are slightly lower. This allow the generation model to synthesize high-motion videos while maintaining overall visual quality.

To achieve these goals, we present \textit{Motion-corrective alignment with Self-critic hierarchical reasoning} (\textit{McSc}),
a three-stage framework comprising two preference reasoning phases and one preference alignment stage. 
In the first phase, we develop a \textit{self-critic dimensional reasoning} (\textit{ScDR}) to train a VLM-based reward model (RM) for per-dimension preference assessment. Using publicly available human-annotated datasets, we decompose multi-dimensional annotations into single-dimension samples and train the RM via rule-based RL~\cite{r1,grpo} to generate both reasoning chains and predictions. The RM self-evaluates the consistency between reasoning and outcomes, providing auxiliary rewards to ensure its internal logic. The ScDR serving as a cold-start phase can establish foundational dimensional reasoning capabilities.
In the second phase, we propose \textit{hierarchical comparative reasoning} (\textit{HCR}) to enable holistic video preference evaluation. The RM learns to generate multi-dimensional structural reasoning outputs for input videos, and then comprehensively analyzes to predict overall preference estimation. We design hierarchical reward mechanisms, including hierarchical format rewards, dimension-wise format rewards, and final comparison preference rewards to boost the RM’s ability. 
For preference alignment in the third stage, we introduce \textit{motion-corrective DPO} (\textit{McDPO}), which leverages the trained RM to generate preference pairs, and then apply a weighted DPO to fine-tune the video generative model. 
To mitigate bias in motion-related dimensions and prevent reward hacking, we compute the motion-related score difference between preferred and disfavored videos in each pair and use it to modulate the optimization process. This correction reduces the influence of biased preference estimation, enabling the model to generate large motion dynamic videos that better align with true human preference.

In summary, our key contributions are outlined below:
\begin{itemize}
\vspace{-2pt}
\item We propose {McSc} for video generation, integrating human preference reasoning and preference alignment to synthesis videos with estimated preference. 
\vspace{-2pt}
\item McSc contains three key steps: 
(1) {ScDR} trains a generative reward model with a self-critic strategy towards single-dimension preference reasoning, 
(2) {HCR} exploits holistic video assessment with structured reward mechanisms, and 
(3) {McDPO} optimizes the video generation model to synthesize diverse videos align with true human preference by reducing evaluation dimension bias.
\vspace{-2pt}
\item Experiments demonstrate that McSc achieves state-of-the-art alignment with human preference on both video preference evaluation and video generation benchmarks.
\vspace{-2pt}
\end{itemize}

%% file: sec/2_related.tex
\section{Related Works}
\label{sec:related}

\subsection{Text-to-Video Generation Model}
\vspace{-2pt}
Text-to-Video (T2V) generation aims to synthesize temporally coherent and semantically aligned videos from natural language descriptions, with modern methods relying on diffusion~\cite{dit} or flow matching~\cite{wan,sora,Cogvideo,Hunyuanvideo,kling,veo3,runway,hailuo,vidu}. Trained on large video-text corpora like OpenVid~\cite{Openvid} and Panda-70M~\cite{Panda}, these models achieves remarkable generalization and can support diverse applications, such as controllable video synthesis~\cite{Dynamicrafter,Make-a-video}, interactive game asset creation~\cite{Gamegen-x}, and robotic simulation~\cite{gr2}.

To improve control and expressiveness, many recent works enhance these models with task-specific modules, for example, DynamicCrafter~\cite{Dynamicrafter} leverages optical flow or motion priors to guide trajectory-aware generation, and OmniAvatar~\cite{OmniAvatar} employs low-rank adaptation (LoRA) to inject audio-driven dynamics into the generation pipeline. Yet most still optimize for pixel-level fidelity over perceptual quality, often producing visually plausible but unnatural or incoherent motion, our focus is to address this gap.

\vspace{-2pt}
\subsection{Preference Alignment for Visual Generation}
\vspace{-2pt}
Recognizing the difficulty of current video generators in producing content that resonates with human judgment, several recent works have turned to reinforcement learning from human feedback (RLHF) to bridge this gap~\cite{imagereward,ddpo,dpok,videoalignment,dpo}. Yet, such methods typically demand large volumes of costly, human-annotated pairwise preference data, which is both time-consuming to collect and difficult to scale. To mitigate this dependency, newer approaches propose automated preference proxies, either through heuristic scoring functions~\cite{ImproveFeedback,Flow-grpo} or vision-language models (VLMs) that assess video-text alignment~\cite{VideoDPO,DanceGRPO,HuViDPO}. While promising, these surrogate signals can be brittle: they may misjudge subtle aspects of visual quality, over-prioritize static frame fidelity, or inadvertently suppress dynamic motion to avoid penalization, resulting in videos that appear “safe” but lack vitality or expressiveness.

In response to these limitations, we introduce a novel preference alignment framework that incorporates explicit preference reasoning. Rather than relying solely on scalar reward signals or simple VLM judgments, our method models the underlying rationale behind human preferences, enabling more robust, interpretable, and motion-preserving alignment without extensive human labeling.

%% file: sec/3_preliminary.tex
\section{Preliminary}

\begin{figure*}[t]
\centering
\includegraphics[width=1.0\textwidth]{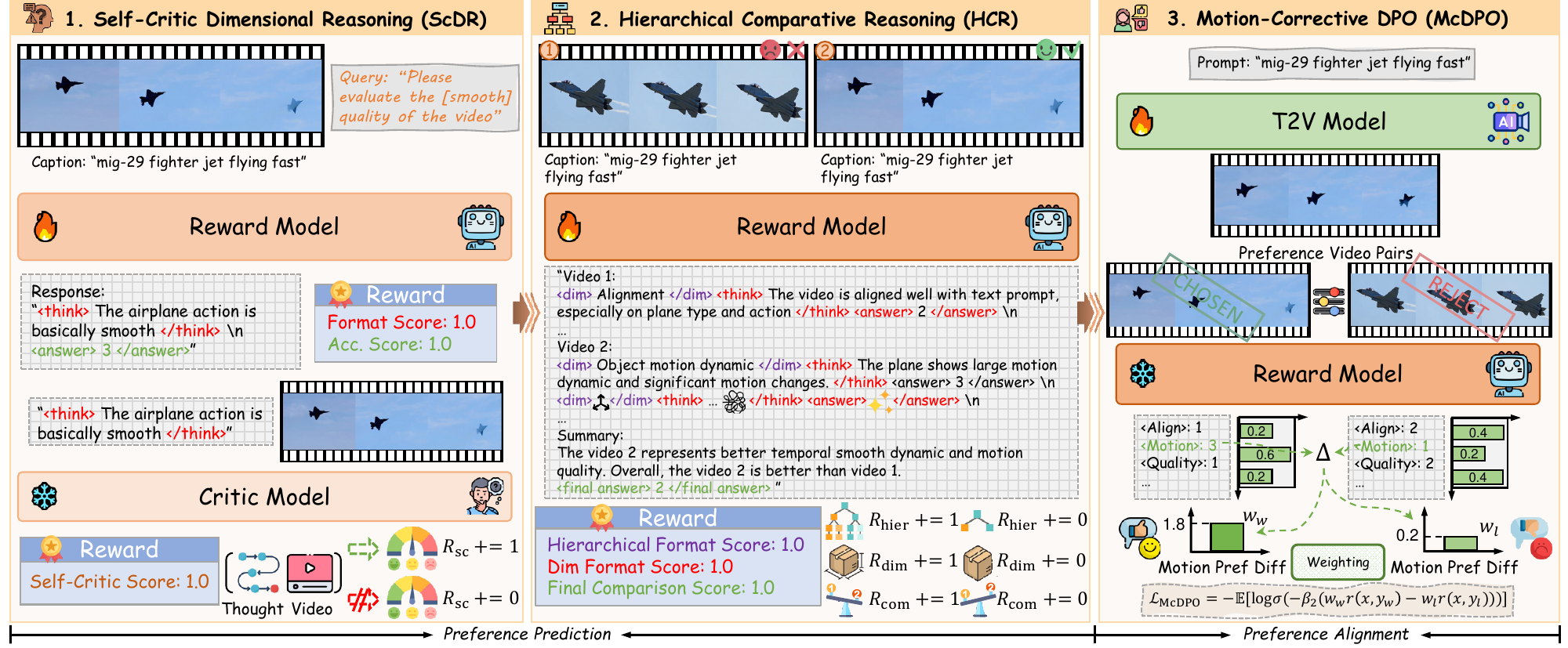} 
\caption{Illustration of the McSc framework containing \textit{preference prediction} and \textit{preference alignment} across three stages. In preference prediction, ScDR as the first phase exploits single-dimension preference judgment, and HCR as the second stage assess overall video quality through multi-dimensional analysis. For preference alignment, McDPO mitigates bias from negatively coupled dimensions for motion-enhanced and reliable alignment.}
\label{fig-framework}
\end{figure*}


\noindent\textbf{Reasoning with Group Relative Policy Optimization (GRPO).}
To incentivize the potential of video preference reasoning, we leverage rule-based RL algorithm, GRPO, to perform reward modeling. 
Given the video and text input $x$, the policy model $\pi_\theta$ samples $G$ reason trajectories to produce diverse responses $\{o_1, o_2,...,o_G\}$. Afterwards, these responses are evaluated under pre-defined rules and assigned reward scores $\{r_1, r_2,...,r_G\}$, which are then normalized and used to calculate the relative advantages $A_i$ within a group for each response $o_i$: $A_i=\frac{r_i-\operatorname{mean}\left(\left\{r_i\right\}_{i=1}^G\right)}{\operatorname{std}\left(\left\{r_i\right\}_{i=1}^G\right)}$.
With these relative advantages, GRPO controls the update magnitude of the policy model with respect to the old policy $\pi_{\theta_{\text{ref}}}$ and maximizes the optimization objective below:
\vspace{-2pt}
\begin{equation}
\label{eq2}
    \begin{aligned}
& \mathcal{L}_{\mathrm{GRPO}}
= \mathbb{E} \bigg[ \frac{1}{G} \sum_{i=1}^G \frac{1}{|o_i|} \sum_{t=1}^{|o_i|}
\min \bigg( \frac{\pi_\theta(o_{i,t}|x,o_{i,<t})}{\pi_{\theta_{\text{ref}}}(o_{i,t}|x,o_{i,<t})} A_{i,t}, \\
&  \operatorname{clip} \left( \frac{\pi_\theta(o_{i,t}|x,o_{i,<t})}{\pi_{\theta_{\text{ref}}}(o_{i,t}|x,o_{i,<t})}, 1-\delta, 1+\delta \right) A_{i,t} \bigg) \bigg],
    \end{aligned}    
\end{equation}
where $\delta$ is the clipping range. We omit a regularization term $-\beta_1 \mathrm{KL}(\pi_\theta||\pi_{\theta_{\text{ref}}})$, where $\mathrm{KL}(\cdot||\cdot)$ denotes the KL divergence and $\beta_1$ refers to the balance factor. 

\noindent\textbf{Preference Alignment with Direct Preference Optimization (DPO).}
DPO is typically employed to align generative models with human preference, which trains on annotated preference pairs with an input condition $(x, v_w, v_l) \in D$, where the positive target $v_w$ is preferred over the negative one $v_l$. The generative model $\pi_\theta$ aims to increase the reward of positive one: $r(x,v_w)=||o_t-\pi_\theta(v_w|x)||_2^2-||o_t-\pi_{\theta_\text{ref}}(v_w|x)||_2^2$, and reduce that of negative one: $r(x,v_l)=||o_t-\pi_\theta(v_l|x)||_2^2-||o_t-\pi_{\theta_\text{ref}}(v_l|x)||_2^2$, where $o_t$ denotes the training target of generation model and $\pi_{\theta_\text{ref}}$ means a frozen reference model initialized by original generation model. The loss function can be written as:
\begin{equation}
    \begin{aligned}
\mathcal{L}_{\mathrm{DPO}}
= & -\mathbb{E}_{D}\left[\text{log} \sigma(-\beta_2 (r(x,v_w)-r(x,v_l)))\right],
    \end{aligned}
\end{equation}
where $\sigma$, $\beta_2$ refer to the normalization and a control coefficient, respectively. 
This approach encourages the model to generate videos that align with the distribution of positive targets, while suppressing the video generation that follow the distribution of negative ones.

%% file: sec/4_method.tex
\section{Methodology}

\noindent\textbf{Overview.}
We present Motion-corrective alignment with Self-critic hierarchical reasoning (McSc) framework (Figure~\ref{fig-framework}), a three-stage training strategy to address both preference prediction (first two stages) and preference alignment (the third stage).
For \textit{preference prediction}, we train a generative reward model (RM) as the predictor using Self-critic Hierarchical Reasoning (ScHR), containing a \textbf{Self-critic Dimensional Reasoning (ScDR)} for per-dimension assessment and a \textbf{Hierarchical Comparative Reasoning (HCR)} for holistic video pair evaluation.
Then, for \textit{preference alignment}, we use the trained RM to yield preferred video pairs and introduce \textbf{Motion-corrective DPO (McDPO)} to align video generation models using video pairs.


\subsection{Self-Critic Dimensional Reasoning (ScDR)}
In the initial stage of our framework, we aim to endow the RM with a strong capability of fine-grained, single-aspect preference reasoning. This stage serves as a cold-start phase, laying the foundation for robust, multi-aspect video preference assessment in subsequent steps.

\noindent\textbf{Preference Data Factorization.}
To supervise RM with precise signals, we first decompose the collected preference-labeled datasets into a series of single-aspect preference instances. Specifically, let $\mathcal{D} = (v_i, y_{i,d})$ denote the training set, where $v_i$ is the $i$-th video, and $y_{i,d}$ means the human-annotated preference label for the $d$-th aspect ($d \in D$). For each $(v, y_d)$ pair, we construct a dimension-specific query $q_d$ explicitly asking about the quality of $v$ for the $d$-th aspect.

\noindent\textbf{Structural Reasoning.}
Each $(v, q_d)$ pair is fed into the RM, which is a pre-trained vision-language model, to obtain a structured response: $o=\text{RM}(v,q_d)$. The model is required to output $o$ in the format: ``\textless think\textgreater reasoning\textless /think\textgreater \textless answer\textgreater prediction\textless /answer \textgreater''. Here, the ``\textless think\textgreater" section contains the step-by-step reasoning process for the $d$-th aspect, while ``\textless answer\textgreater" presents the quality prediction for that aspect.
To robustly assess the model’s output, we sample the response $o$ multiple times for each $(v, q_d)$. 
For each sampled output, we compute two reward scores.
The first is Format Score ($r_{\mathrm{format}}$), which indicates whether $o$ adheres to the required response format. $r_{\mathrm{format}} = 1$ if the format is correct, otherwise $0$.
The second is Accuracy Score ($r_{\mathrm{acc}}$) that measures whether the predicted label in ``\textless answer\textgreater" matches the ground-truth $y_d$. $r_{\mathrm{acc}} = 1$ if correct; $0$ otherwise.

\noindent\textbf{Self-Critic Consistency Evaluation.}
Beyond format and prediction accuracy, we introduce a self-critic mechanism to assess the reasoning consistency of the RM. Obtained $o$ from RM, we treat current RM as a critic model (CM) and prompt it with $o$ and the video as input. The CM evaluates whether the reasoning in ``\textless think\textgreater \textless /think\textgreater" logically supports the prediction in ``\textless answer\textgreater \textless /answer\textgreater" according to the video. If the self-critique returns a positive assessment, we assign a self-critic reward score $r_{\mathrm{sc}}=1$; otherwise $0$.

\noindent\textbf{Reward Aggregation and Model Optimization.}
We combine the three reward scores into a final reward:
\begin{equation}
    r=r_\text{format}+r_\text{acc}+r_\text{sc}.
\end{equation}
The RM is updated via GRPO by maximizing the expected reward over the dataset as Eq.~(\ref{eq2}).
In this manner, RM is guided to understand and internalize human-like reasoning processes for each aspect of video quality with logically support, which can be maintained justification consistency via self-critic mechanism.
This stage serves as a cold-start initialization, ensuring the RM to perform robust single-aspect preference reasoning, representing a prerequisite for more holistic and comparative assessments in the next stage.

\subsection{Hierarchical Comparative Reasoning (HCR)}
Building upon the RM’s initial single-aspect reasoning ability, we further develop its capacity to conduct comprehensive and hierarchical comparative reasoning for robust video quality evaluation. This stage enables the RM to not only assess videos from multiple individual dimensions but also comprehensively compare the overall preference between two videos.

\noindent\textbf{Hierarchically Structured Evaluation.}
Given a pair of videos $(v_a, v_b)$, we prompt the model with a query $q_{\mathrm{hier}}$ designed to elicit a hierarchically structured response across diverse dimensions. 
This structure enforces the RM to independently reason and predict for each aspect $d = 1,2,3...$, then aggregate these aspect-level judgments to produce a final overall quality comparative prediction.

\noindent\textbf{Hierarchical Rewards.}
To ensure the model learning to perform well-structured and consistent reasoning, we introduce three rewards at this stage:

\begin{itemize}

\item \textit{Hierarchy Format Score} ($r_{\mathrm{hier}}$):
$r_{\mathrm{hier}} = 1$ if response $o_v$ contains diverse evaluation dimensions in ``\textless dim\textgreater \textless /dim\textgreater" format, $0$ otherwise.

\item \textit{Dimension Format Score} ($r_{\mathrm{dim}}$):
For each aspect $d$, $r_{\mathrm{dim}, d} = 1$ if the output for aspect $d$ matches the required format, $0$ otherwise. The overall dimension format score is averaged across aspects: $r_{\operatorname{dim}}=\frac{1}{D} \sum_{d=1}^D r_{\operatorname{dim}, d}$.

\item \textit{Final Comparison Score} ($r_{\mathrm{com}}$):
After obtaining $o_{v_a}$ and $o_{v_b}$, we compare their respective overall quality predictions and reasoning. The model is then prompted to output a final preference prediction $y^* \in {a, b}$, indicating which video is preferred. If $y^*$ matches the ground-truth human label $y_{\mathrm{gt}}$, $r_{\mathrm{com}} = 1$, otherwise $0$.

\end{itemize}

\noindent\textbf{Reward Aggregation and Training.}
The total reward for each video pair is defined as:
\begin{equation}
    r_\text{total}=r_\text{hier}+r_\text{dim}+r_\text{com}.
\end{equation}
With this total reward modeling, the RM are optimized to maximize the expected total reward via the GRPO algorithm as described in Eq.~(\ref{eq2}).
This hierarchical reasoning process inherits the single-aspect reasoning skills acquired during the first cold-start phase, while learning to synthesize multi-aspect judgments for holistic video quality assessment. By enforcing structured output and logical aggregation, the RM develops reliable overall preference estimation between pairs of videos.

\textit{The ScDR and HCR constitute a two-stage RM training scheme, ScHR, for preference prediction.} ScHR enables the RM to automatically annotate preference labels for video pairs, thereby ScHR substantially reducing the reliance on manual annotation while providing high-quality supervision for subsequent video generation model alignment.

\subsection{Motion-Corrective DPO (McDPO)}
Obtained the RM trained by ScHR, we leverage it to produce pseudo preference labels of all videos for aligning the video generation model $\pi_\theta$ with human preference. 
Our preliminary analysis reveals a significant negative correlation between motion-related dimensions and other quality scores produced by RM, causing a bias that suppresses the video motion dynamic degree. 
To tackle this, we propose McDPO to rectify this bias and enable a more authentic alignment with human preference for high-motion dynamic content.

\noindent\textbf{Constructing Preference Video Pairs.}
Given a text prompt $x$, we sample a set of $N$ candidate videos ${v_1, v_2, \dots, v_N}$ from pre-trained video generation model $\pi_\theta$. For each generated video $v_i$, the RM is used to estimate an overall preference score $s_i=\text{RM}(v_i)$, where $s_i$ reflects the estimated degree to which $v_i$ aligns with human preference, as inferred by the RM.
Afterwards, to construct effective training pairs for preference alignment, we identify the video with the highest preference score as the positive example: $v_w=\arg \max_{v_i} s_i$ and the one with the lowest score as the negative example: $v_l=\arg \min_{v_i} s_i$. This selection ensures a clear preference margin for learning.

\noindent\textbf{Motion-Corrective Weighting.}
Considering the remarkable negative correlation between motion-related aspects (e.g., object and camera motion dynamic) and other dimensions of visual quality, we identify a bias that leads the model to prefer videos with low motion dynamic during preference alignment.
To mitigate this, we propose a motion-corrective re-weighting strategy for the DPO objective. Specifically, we compute correction weights based on the object motion dynamic scores $s_w^\text{om}, s_l^\text{om}$ and camera motion dynamic scores $s_w^\text{cm}, s_l^\text{cm}$ of the positive (win) and negative (lose) videos, respectively, predicted by the RM. Afterwards, we calculate the re-weighting terms $w_w, w_l$ for win and lose videos as:
\begin{equation}
\begin{aligned}
    w_w &= 0.5 + \sigma\left[(s_w^\text{om} - s_l^\text{om}) + (s_w^\text{cm} - s_l^\text{cm})\right], \\
    w_l &= 2.0 - w_w,
\end{aligned}
\end{equation}
where $\sigma$ means sigmoid function as the normalization for evaluation scores.
These weights reflect the potential influence of motion-related quality on overall video preference.



\begin{table}[t]
\renewcommand\arraystretch{1}
\setlength{\tabcolsep}{2pt}  
\centering
\small
\begin{tabular}{lccccc}
\toprule
\multirow{2}{*}{\textbf{Methods}} & \multirow{2}{*}{\textbf{Reason}} & \multicolumn{2}{c}{\textbf{MonetBench}} & \multicolumn{2}{c}{\textbf{GenAI-Bench}} \\ \cline{3-6} 
\multicolumn{1}{c}{} && \textbf{tau} & \textbf{diff} & \textbf{tau} & \textbf{diff} \\ \hline
GPT-4o~\cite{gpt4}        & $\times$ & 45.7 & 48.3 & 41.8 & 54.3 \\ 
Genimi~\cite{Gemini}        & $\times$ & 52.2 & 56.8 & 46.9 & 61.7 \\ 
VQAScore~\cite{vqascore}      & $\times$ & 49.1 & 54.9 & 45.2 & 68.0 \\ 
VideoScore~\cite{Videoscore}    & $\times$ & 49.1 & 54.9 & 47.8 & 71.4 \\ 
VisionReward~\cite{Visionreward}  & $\times$ & 64.0 & 72.1 & 51.8 & 74.4 \\ 
VideoReward~\cite{videoalignment}   & $\times$ & - & - & 50.2 & 73.3 \\ 
UnifiedReward~\cite{unifiedreward} & $\times$ & - & - & 60.7 & 77.2 \\
UnifiedReward-CoT~\cite{unifiedreward-think} & $\checkmark$ & - & - & - & 82.3 \\ \hline
ScHR w/o Reasoning & $\times$  & 68.1 & 77.5 & 57.4 & 77.7 \\ 
{ScHR w/o ScDR}    & $\checkmark$ & 70.7 & 79.5 & 60.6 & 80.8 \\ 
{ScHR w/o HCR}     & $\checkmark$ & 69.5 & 78.1 & 59.3 & 79.2 \\ 
\textbf{ScHR}      & $\checkmark$ & \textbf{72.0} & \textbf{81.5} & \textbf{62.9} & \textbf{82.7} \\ \midrule
\end{tabular}
\caption{Preference accuracy on human video preference evaluation benchmarks. \textit{tau} denotes taking account of ties and \textit{diff} represents dropping ties in labels.}
\label{table-preference}
\end{table}

\noindent\textbf{Motion-Corrective Preference Alignment.}
For each text prompt $x$ and generated videos $v_w,v_l$, the McDPO loss for the tuple is defined as:
\begin{equation}
    \begin{aligned}
\mathcal{L}_{\mathrm{McDPO}}
= & -\mathbb{E}_{D}\left[\text{log} \sigma(-\beta_2 ({w}_w  r(x,v_w)-{w}_l r(x,v_l)))\right].
    \end{aligned}
\end{equation}
These weights reflect the extent to which motion dynamic influence the overall preference prediction. If the positive video exhibits low-motion dynamic scores, its weight will be reduced to mitigate the bias introduced by the negative coupling between motion and other quality dimensions. By minimizing the McDPO objective, the generative model increases the likelihood of producing videos that align with the RM’s assessment (i.e., the estimated human preference), while mitigating spurious correlations that may incur reward hacking. This results in more robust and faithful alignment with real human judgment.

%% file: sec/5_exp.tex
\section{Experiments}

\subsection{Preference Reasoning}

\noindent\textbf{Training Datasets.} 
We collect 121.1k preference-annotated videos from five public preference datasets, including VisionReward~\cite{Visionreward}, MJ-Video~\cite{Mj-video}, Lift-HRA~\cite{Lift}, VideoDPO~\cite{VideoDPO}, and Charades~\cite{Charades}. These datasets contain diverse annotation types, incuding scoring or ranking, across various dimensions, resulting in totally 679.5k dimensional assessment labels.

\begin{table}[t]
\renewcommand\arraystretch{1}
\setlength{\tabcolsep}{2.3pt}
\centering
\small
\begin{tabular}{lcccc}
\toprule
\multirow{2}{*}{\textbf{Methods}} & \multirow{2}{*}{\textbf{Preference}} & \multicolumn{3}{c}{\textbf{VBench}}  \\ \cline{3-5} 
\multicolumn{1}{c}{} && \textbf{Total} & \textbf{Quality} & \textbf{Semantic}  \\ \hline
{\textbf{\textit{VideoCrafter2}}} &&&& \\
Baseline~\cite{vc2} & - & 80.44 & 82.20 & 73.43  \\ 
SFT      & - & 78.47 & 79.29 & 74.11  \\ 
VADER~\cite{videoalignment}    & VADER-Reward & 80.59 & 82.46 & 73.09  \\ 
VideoDPO~\cite{VideoDPO} & OmniScore    & 81.93 & 83.07 & 77.38  \\ 
Flow-DPO~\cite{ImproveFeedback} & VideoReward  & 81.35 & 83.35 & 77.04  \\ 
\textbf{McSc}  & \textbf{ScHR} & \textbf{83.97} & \textbf{85.03} & \textbf{80.77}  \\ \hline



{\textbf{\textit{Wan2.1-T2V-1.3B}}} &&&& \\
Baseline~\cite{wan} & - & 83.96 & 84.92 & 80.10 \\ 
SFT      & - & 84.20 & 84.67 & 80.46  \\ 
VideoDPO~\cite{VideoDPO} & OmniScore& 84.82 & 85.33 & 80.85  \\ 
Flow-DPO~\cite{ImproveFeedback} & VideoReward  & 85.07 & 85.05 & 80.94  \\ 
\textbf{McSc} & \textbf{ScHR} & \textbf{85.71} & \textbf{85.77} & \textbf{81.84}   \\  \midrule
\end{tabular}
\caption{Video generation alignment performance on VBench guided by different optimization approach and preference estimation manners.}
\label{table-video-gen}
\end{table}

\begin{figure*}[t]
\centering
\includegraphics[width=0.99\textwidth]{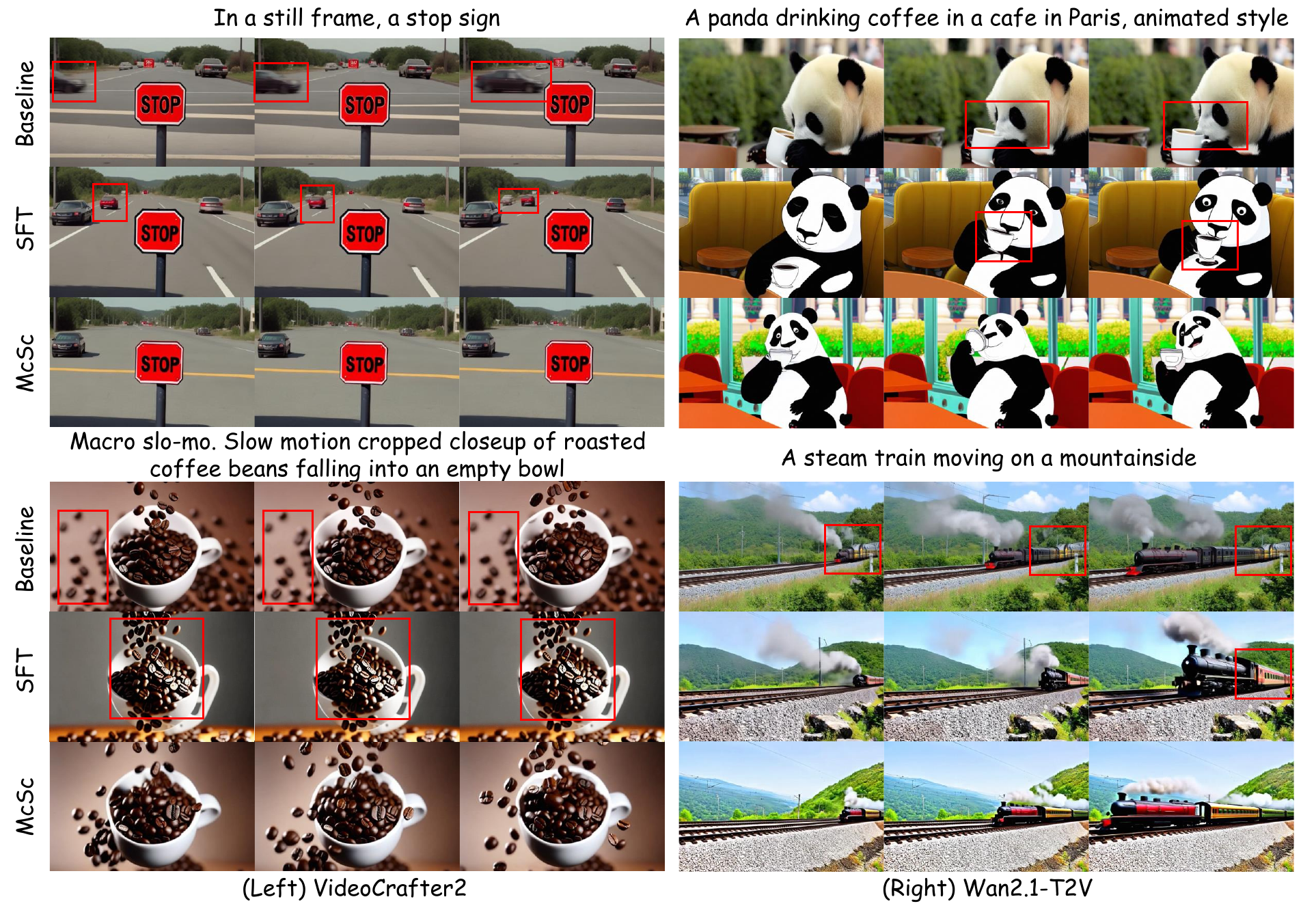}
\caption{Visualization comparison of generated videos by the baselines and the proposed method. Our McSc generate videos with larger motion dynamic and stronger semantic alignment.}
\label{fig-comparison}
\end{figure*}

\noindent\textbf{Evaluation.}
To evaluate our reward model on preference prediction, we use two benchmarks: (1) GenAI-Bench~\cite{Genai-bench} contains short videos from prior T2V models; (2) MonetBench~\cite{Visionreward} includes 13,342 high-quality prompts from VidProM~\cite{Vidprom}, covering diverse category annotations for fine-grained assessment.

\noindent\textbf{Implementation Details.}
We initialize the reward model and critic model adopting a pre-trained Qwen2-VL-7B-Instruct~\cite{qwenvl} model, and efficiently fine-tune the reward model via LoRA~\cite{lora}. During the whole preference reasoning, the reward model is updated by the AdamW~\cite{adamw} with an initial learning rate of $1\times10^{-6}$ via a linear decay, rollout number $G$ of 8, $\beta_1$ of 0.07, batch size of 16 and epoch of 2.

\noindent\textbf{Quantitative Results.}
As suggested in Table~\ref{table-preference}, ScHR achieves 71.9\%/81.2\% preference prediction accuracy on MonetBench and 62.4\%/81.8\% on GenAI-Bench under \textit{tau} and \textit{diff} settings, respectively, outperforming all comparisons~\cite{gpt4,Gemini,Visionreward,ImproveFeedback,unifiedreward,unifiedreward-think}. Notably, on GenAI-Bench, ScHR surpasses the best non-reasoning model UnifiedReward~\cite{unifiedreward} by 1.7\% and 4.6\%, highlighting the clear benefits of incorporating reasoning capabilities in predicting human preference for videos. Compared to the latest reasoning-based method UnifiedReward-CoT~\cite{unifiedreward-think}, which directly applies GRPO to preference reasoning, our ScHR exceeds it despite using a smaller model, demonstrating the effectiveness of our single-dimension reasoning then hierarchical reasoning scheme on capturing complex human judgment patterns.

\noindent\textbf{Ablation Study.}  
To evaluate ScHR and its components (ScDR and HCR), we conduct ablation studies on the preference prediction task.
\noindent\textbf{(1) Affect of ScDR.}
Removing ScDR and using only HCR for direct overall preference prediction (ScHR w/o ScDR in Table~\ref{table-preference}) reduces results by 1.3\% in \textit{tau} across both benchmarks, indicating that ScDR as a cold-start phase enables the model to capture per-dimension judgment logic for better estimation.
\noindent\textbf{(2) Influence of HCR.}
Without HCR, average aggregating ScDR’s per-dimension scores yields significantly worse results (ScHR w/o HCR in Table~\ref{table-preference}), confirming HCR’s crucial role in holistic quality assessment.
\noindent\textbf{(3) Analysis of Reasoning.} 
Figure~\ref{fig-motivation} (c) illustrates the model’s logical reasoning process. Disabling reasoning rewarding and training directly on preference labels (ScHR w/o Reasoning) causes a notable drop, demonstrating that explicit reasoning can enhance prediction accuracy.
\noindent\textbf{(4) Discussion on Computation Cost.}
Our two-stage strategy strikes an effective trade-off balance: it enhances preference prediction through lightweight RL optimization without increasing inference cost, enabling efficient, manually annotation-free preference alignment.

\vspace{-2pt}
\subsection{Video Generation with Preference Alignment}
\vspace{-2pt}
\noindent\textbf{Training Datasets.}
We construct tailored video datasets for different generation models to ensure effective DPO optimization. For VideoCrafter2, we re-annotate VideoDPO dataset using our learned reward model to generate preference labels. For Wan2.1-T2V-1.3B, we generate four videos per text prompt from the VideoDPO dataset using the pre-trained model, and then employ our reward model to annotate preference and construct video pairs with win/loss labels, serving as the training data.

\begin{figure}[t]
\centering
\includegraphics[width=0.49\textwidth]{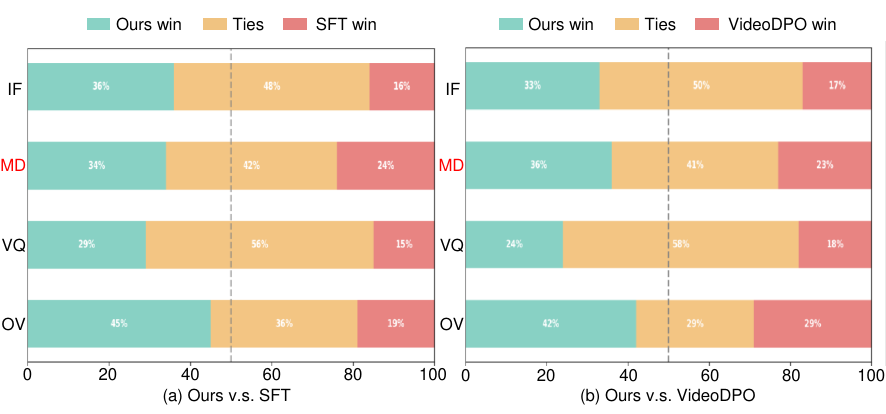} 
\caption{Human evaluation on preference rate of our model with SFT and VideoDPO on VBench across four aspects, including instruction-following (IF), motion dynamic (MD), visual quality (VQ) and overall (OV) performance.}
\label{fig-win-loss}
\end{figure}

\begin{figure}[t]
\centering
\includegraphics[width=0.48\textwidth]{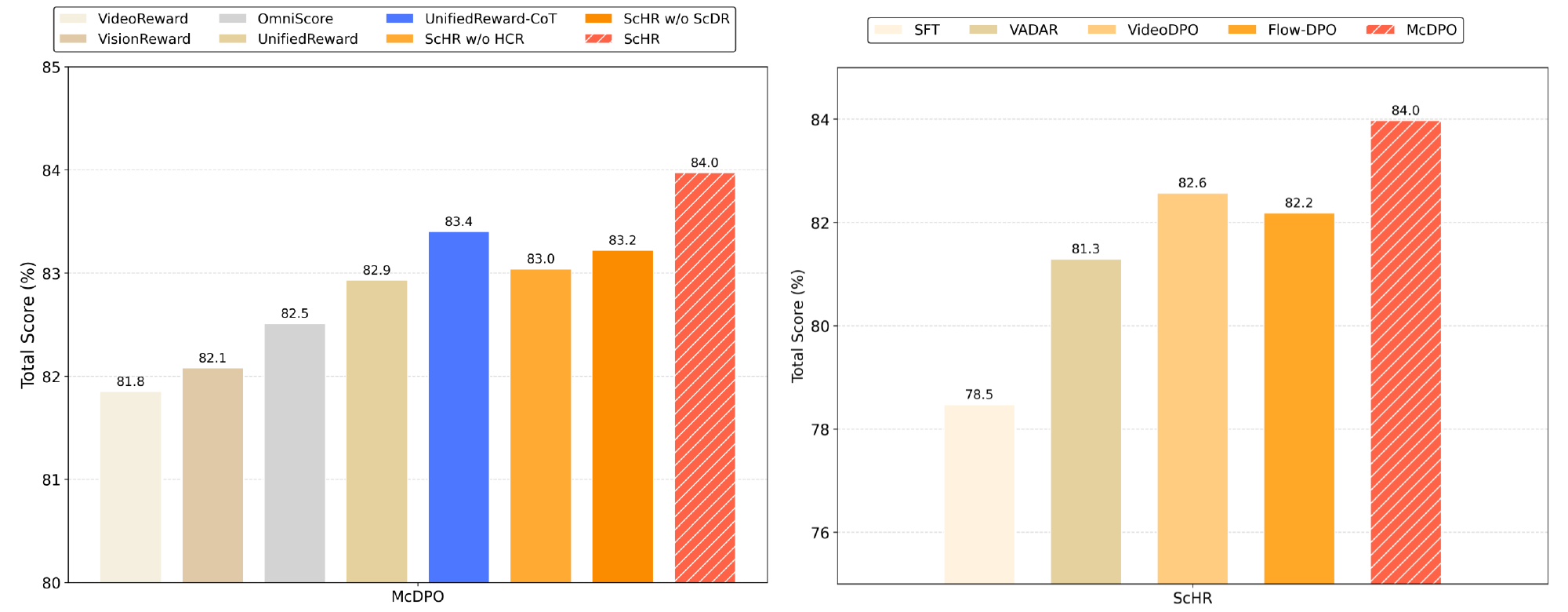}
\caption{Ablation study and comparison with alternative approaches for ScHR (left) and McDPO (right) on VBench.}
\label{fig-ablation-com}
\vspace{-2pt}
\end{figure}

\noindent\textbf{Implementation Details.}
We initialize the DPO fine-tuning process using pre-trained versions of both VideoCrafter2~\cite{vc2} and Wan2.1-T2V-1.3B~\cite{wan} as baseline video generation models. Upon the constructed preference pairs, we select the winning videos and perform supervised fine-tuning (SFT) on baseline models as the SFT models, and utilize 10k winning and lose paired videos to optimize models via our McDPO alignment strategy. The video models are trained by AdamW optimizer with an initial learning rate of $6\times 10^{-6}$, $\beta_2$ of 2,500, batch size of 8 and total epoch of 20, using 8 NVIDIA A100 (80G) GPUs.

\noindent\textbf{Evaluation.}
To assess the quality of generated videos, we employ VBench~\cite{vbench} as the benchmark, which contains specific text prompts and various evaluation dimensions.

\noindent\textbf{Quantitative Results.}
We evaluate McSc by comparing the full framework and its individual components against the latest approaches on VBench.
\noindent\textbf{(1) Overall Comparison.}
From Table~\ref{table-video-gen}, using VideoCrafter2 as the base model, our method achieves scores of 83.97\%, 85.03\% and 80.77\% in total quality, visual quality, and semantic alignment, respectively. With Wan2.1 as the backbone, we achieve 85.71\%, 85.77\% and 81.84\%, significantly outperforming both the pre-trained baseline and the SFT model. These results also surpass recent alignment methods~\cite{VideoDPO,videoalignment,ImproveFeedback}, verifying McSc's strongest video generation and alignment capability.
\noindent\textbf{(2) Comparison for Preference Prediction and Alignment.}
We compare ScHR and McDPO with latest methods on preference prediction and alignment tasks, respectively. For prediction, we fix McDPO and compare generation performance using data from different predictors. For alignment, we fix ScHR and evaluate various alignment strategies. Figure~\ref{fig-ablation-com} shows our methods achieve superior performance in both tasks, demonstrating their advantages.

\noindent\textbf{Qualitative Results.}
We provide qualitative comparisons through synthesis videos and human evaluation.
\noindent\textbf{(1) Visualization of Generated Videos.}
As shown in Figure~\ref{fig-comparison}, the baseline and SFT models often produce prompt-misaligned or artifact-laden videos with low motion dynamic, whereas McSc generates more realistic and prompt-compliant results with high motion dynamic, highlighting the effectiveness of our alignment strategy.
\noindent\textbf{(2) Human Evaluation.}
To better assess video quality and preference alignment, we conduct human evaluation on 100 groups of generated videos. From Figure~\ref{fig-win-loss}, McSc outperforms others in instruction following, motion quality, visual quality, and overall preference, demonstrating superior alignment with real human preference and high-quality video generation.

\begin{table}[t]
\renewcommand\arraystretch{1}
\setlength{\tabcolsep}{4.6pt}
\centering
\small
\begin{tabular}{l|cc|cccc}
\toprule
\multirow{2}{*}{\textbf{Methods}} & \multicolumn{2}{c|}{\textbf{McDPO}} & \multirow{2}{*}{\textbf{DD}} & \multirow{2}{*}{\textbf{MS}} & \multirow{2}{*}{\textbf{SC}} & \multirow{2}{*}{\textbf{Total}}  \\ 
& \textbf{OM} & \textbf{CM} & &&& \\ \midrule
Baseline & - & - & 42.50 & 97.73 & 96.85 & 80.44 \\ 
SFT      & - & - & 40.06 $\downarrow$ & 96.15 $\downarrow$ & 95.90 $\downarrow$ & 78.47 \\ 
VideoDPO & - & - & 32.64 $\downarrow$ & 92.18 $\downarrow$ & 95.69 $\downarrow$ & 81.93 \\ \hline
\multirow{3}{*}{{Ours}} & $\checkmark$ & - & {40.57} $\uparrow$ & {96.07} $\uparrow$ & {96.04} $\uparrow$  & {83.15}\\
& - & $\checkmark$              & {38.65} $\uparrow$ & {95.33} $\uparrow$ & {96.79} $\uparrow$  & {82.81}\\ 
& $\checkmark$ & $\checkmark$   & \textbf{43.26} $\uparrow$ & \textbf{97.88} $\uparrow$ & \textbf{97.24} $\uparrow$  & \textbf{83.97}\\ \bottomrule
\end{tabular}
\caption{Analysis of the correctness on motion dynamic by McDPO on VBench. OM and CM represent object and camera motion dynamic. DD, MS, SC denote dynamic degree, motion smooth, subject consistency, respectively. SFT and VideoDPO suggest drops on DD ($\downarrow$) while ours yield increases ($\uparrow$) over VideoDPO.}
\label{table-ablation}
\vspace{-0.2cm}
\end{table}

\noindent\textbf{Ablation Study.}
We investigate the effectiveness on video generation for each proposed component using VideoCrafter2 as the base model.
\noindent\textbf{(1) Effect of ScHR.}
To evaluate the effect of ScHR (with ScDR and HCR) on video generation quality, we adopt various recent preference prediction approaches to generate video pairs and optimize the model via McDPO, followed by evaluation on VBench. As suggested in Figure~\ref{fig-ablation-com} (left), ScHR outperforms recent baselines, demonstrating its effectiveness in generating more alignable preference outcomes.
\noindent\textbf{(2) Influence of McDPO.}
To compare McDPO with recent alignment methods, we fix the preference data from ScHR and apply different alignment algorithms. Figure~\ref{fig-ablation-com} (right) shows that McDPO achieves the best generation performance, confirming its superiority in preference alignment.
\noindent\textbf{(3) Analysis of Motion Correctness.}
We analyze McDPO’s ability to address negative correlation between motion-related and other dimensions, which biases models toward low-motion outputs. From Table~\ref{table-ablation}, progressively re-weighting object (OM) and camera motion (CM) correctness improves dynamic degree (DD) over VideoDPO, while preserving overall quality. Figure~\ref{fig-comparison} further validates enhanced motion dynamics, demonstrating that McDPO can mitigate reward hacking from dimension coupling without sacrificing video quality.

%% file: sec/6_conclusion.tex
\section{Conclusion}

We introduce McSc, which consists of preference prediction and alignment across three training phases. During preference prediction, we present ScHR, including ScDR and HCR stages, for reliable hierarchical reasoning. For preference alignment, McDPO mitigates the motion-guided evaluation bias for improved video generation with high motion dynamic.
Experiments verify that McSc can synthesize high-quality videos aligning more closely with human preference and exhibits diverse dynamic motion.